\pdfoutput=1
\documentclass{bmvc2k}

\usepackage{times}
\usepackage{epsfig}
\usepackage{graphicx}
\usepackage{amsmath}
\usepackage{amssymb}
\usepackage{wrapfig}

\usepackage{colortbl}
\usepackage{xspace}
\definecolor{Gray}{rgb}{0.95,0.95,0.95}
\usepackage{multirow}
\usepackage{adjustbox}
\usepackage{array}
\usepackage{caption}
\usepackage{subcaption}

\usepackage{enumitem,kantlipsum}

\newcommand{\cqd}{CQD\xspace}
\newcommand{\cA}{${\cal A}$\xspace}
\newcommand{\cB}{${\cal B}$\xspace}

\newcommand{\cAB}{${\cal A} + {\cal B}$\xspace}
\newcommand{\cAsB}{${\cal A} , {\cal B}$\xspace}

\title{Adapting Models to Signal Degradation using Distillation}

\addauthor{Jong-Chyi Su}{jcsu@cs.umass.edu}{1}
\addauthor{Subhransu Maji}{smaji@cs.umass.edu}{1}

\addinstitution{
 University of Massachusetts, Amherst\\
 Amherst, MA 01003
}

\runninghead{Su, Maji}{Adapting Models to Signal Degradation using Distillation}

\def\eg{\emph{e.g}\bmvaOneDot}

\def\etal{\emph{et al}\bmvaOneDot}

\begin{document}
\hoffset=0.15in
\maketitle

\begin{abstract}
Model compression and knowledge distillation have been successfully applied for cross-architecture and cross-domain transfer learning. However, a key requirement is that training examples are in correspondence across the domains. We show that in many scenarios of practical importance such aligned data can be synthetically generated using computer graphics pipelines allowing domain adaptation through distillation. We apply this technique to learn models for recognizing low-resolution images using labeled high-resolution images, non-localized objects using labeled localized objects, line-drawings using labeled color images, etc. Experiments on various fine-grained recognition datasets demonstrate that the technique improves recognition performance on the low-quality data and beats strong baselines for domain adaptation. Finally, we present insights into workings of the technique through visualizations and relating it to existing literature.
\end{abstract}

\section{Introduction}
One of the challenges in computer vision is to build models for recognition that are robust to various forms of degradation of the quality of the signal such as loss in resolution, lower signal-to-noise ratio, poor alignment of the objects in images, etc. For example, the performance of existing models for fine-grained recognition drop rapidly when the resolution of the input image is reduced (see Table~\ref{tab:main-results}). 

\begin{wrapfigure}{r}{0.47\textwidth}
\vspace{-0.2in}
\centering
\includegraphics[width=0.9\linewidth]{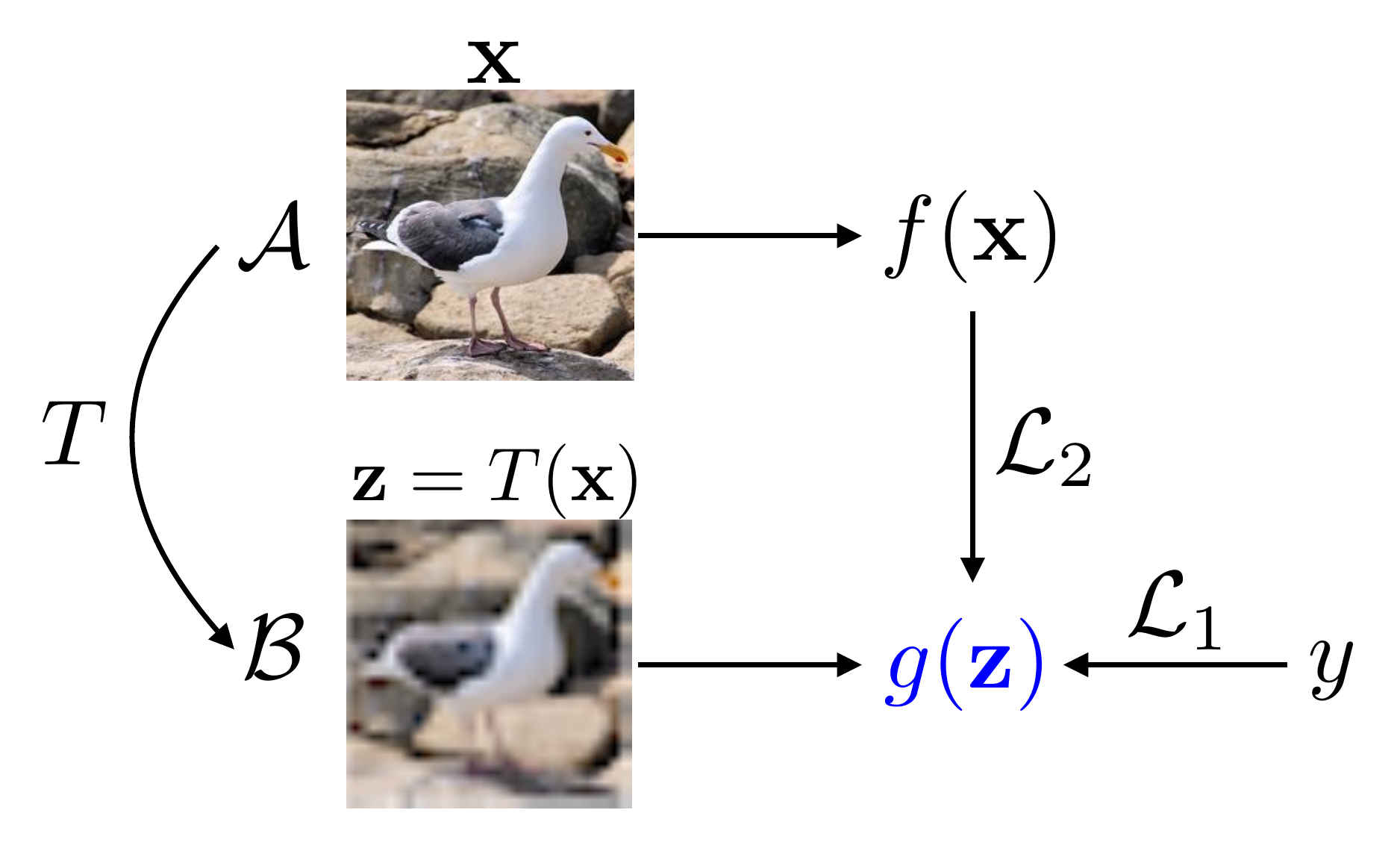} 
\caption{
\label{fig:cqd-framework} The objective of the CQD encourages agreement between $g(z)$ and $f(x)$ for each $z=T(x)$.
}
\vspace{-0.2in}
\end{wrapfigure}

In many cases abundant high-quality data is available at training time, but not at test time. For example, one might have high-resolution images of birds taken by a professional photographer, while an average user might upload blurry images taken from their mobile devices for recognition. We propose a simple and effective way of adapting models in such scenarios. The idea is to \emph{synthetically} generate data of the second domain from the first and \emph{forcing agreement} between the model predictions across domains (Figure~\ref{fig:cqd-framework}). The approach is a simple generalization of a technique called model compression, or knowledge distillation~\cite{Caruana06,Hinton14,ba2014deep}.  

\restoregeometry
The main contribution of our work is to identify several practical scenarios where this idea can be applied. The simplest case is when domain ${\cal B}$ is a ``degraded'' version of domain ${\cal A}$. For example, when domain ${\cal B}$ has lower resolution than ${\cal A}$, or has no color information. It is easy to generate aligned data by applying known transformations $T$ to obtain paired data of the form $[\mathbf{x}, T(\mathbf{x})]$. We also identify some non-trivial cases, such as when domain ${\cal A}$ has images of objects with known bounding-boxes while domain ${\cal B}$ does not. In such situations, a common approach is to train an object detector to localize the object and then classify the image. Our approach offers an alternate strategy where we first train a model on the cropped images and distill it to a model on full images. Experiments show that the improvements are significant, and in some cases matching the results using an object detector. Similarly, we can apply our technique to recognize distorted images as an alternative to Spatial Transformer Networks~\cite{jaderberg2015spatial}. We call our approach Cross Quality Distillation (CQD).

We perform experiments on recognizing fine-grained categories of birds and cars using off-the-shelf Convolutional Neural Networks (CNNs). 
Experiments are performed on improving the recognition of low-quality data using high-quality data with various kinds of degradation (Figure~\ref{fig:dataset}).
This is a challenging task even on the high-quality images, but performance of the models are often dramatically lower when directly applied on the low-quality images. Our experiments show that \cqd leads to significant improvements over a model trained directly on the low-quality data and other strong baselines for domain adaptation, such as fine-tuning and ``staged training''~\cite{Saenko_ICIP16}. 
The model works across a variety of tasks and domains without any task-specific customization. Finally, we present insights into why the method works by relating it to the area of curriculum learning~\cite{bengio2009curriculum} and through visualizations of the learned models.

\section{Related Work}
\label{sec:related}
\begin{figure*}[t]
\hspace*{-0.25cm}
\centering
\begin{tabular}{cccc}
\includegraphics[height=0.12\linewidth]{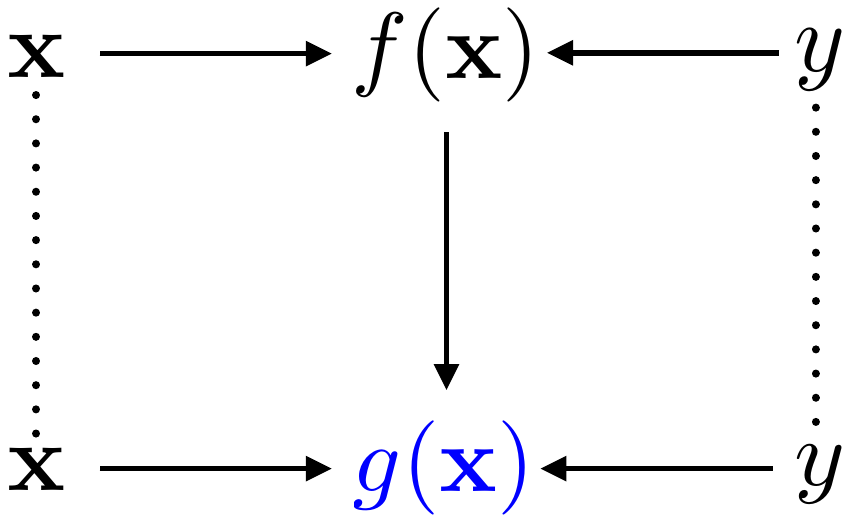} & 
\includegraphics[height=0.12\linewidth]{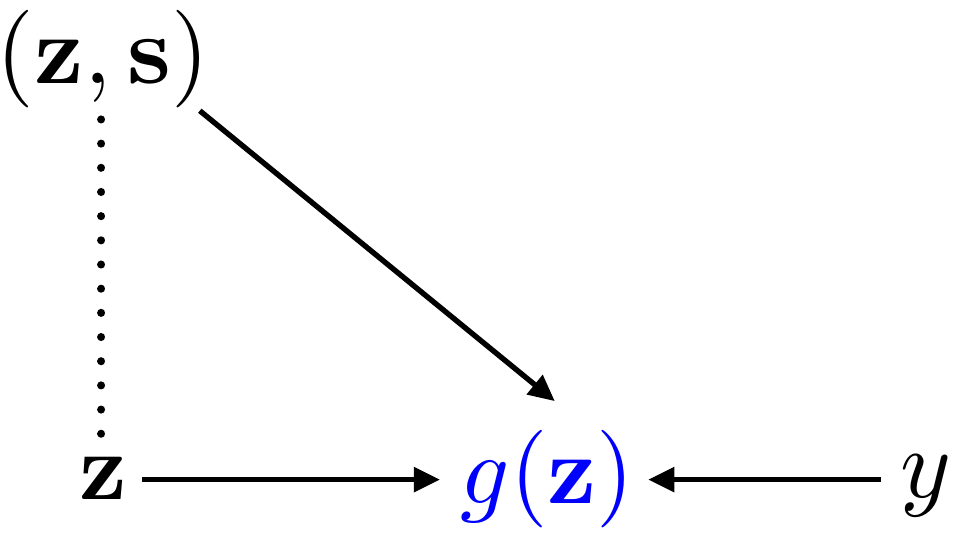} & 
\includegraphics[height=0.12\linewidth]{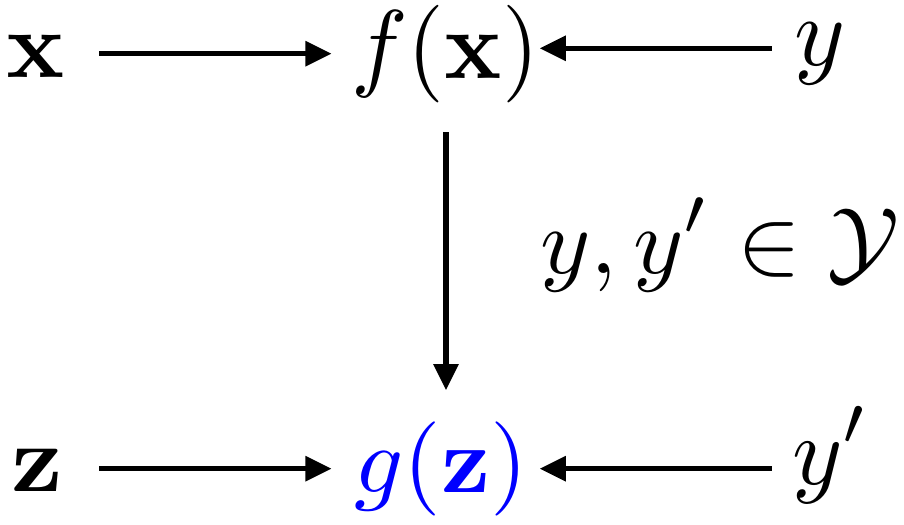} & 
\includegraphics[height=0.12\linewidth]{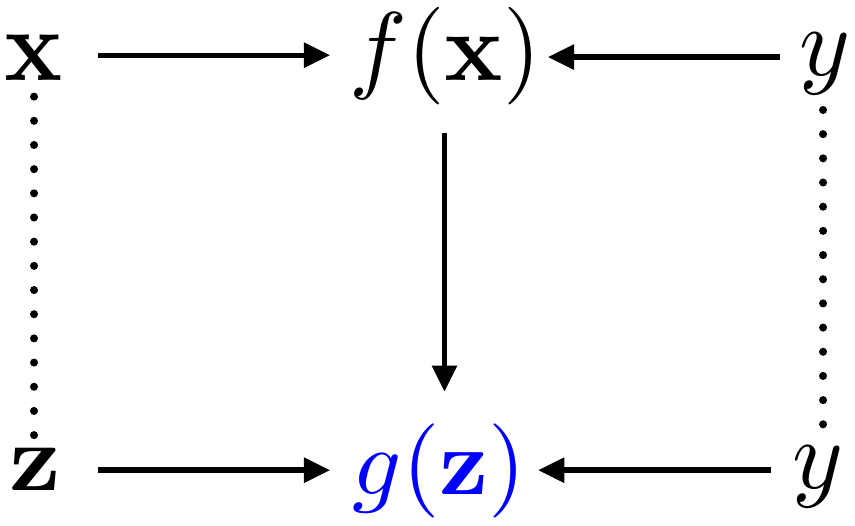}\\
(a) model compression &
(b) LUPI &
(c) domain adaptation &
(d) \cqd
\end{tabular}
\vspace{0.15in}
\caption{
\label{fig:cqd-compare}
Illustration of the relationships between \cqd and other techniques. 
An arrow points to the direction of variable dependency, and dotted lines denote that the variables are observed together. 
\emph{(a) Model compression:} $g$ is trained to mimic the outputs of $f$ on the same input ${\bf x}$.
\emph{(b) LUPI:} $g$ is trained with side information ${\bf s}$ observed together with input ${\bf z}$.
\emph{(c) Domain adaptation:} ${\bf x}$ and ${\bf z}$ are drawn independently from different domains but the tasks are the same, i.e. $y, y' \in \mathcal{Y}$. 
\emph{(d) \cqd:} can be seen as (i) a generalization of model compression where the inputs of the two functions are different, (ii) a specialization of domain adaptation when ${\bf z}$ can be synthesized from ${\bf x}$.
}
\vspace{-0.2in}
\end{figure*}

\paragraph{Knowledge distillation}
The proposed approach is inspired by ``knowledge distillation'' technique~\cite{Hinton14} where a simple classifier $g$, \eg a shallow neural network, is trained to imitate the outputs of a complex classifier $f$, \eg a deep neural network (Figure~\ref{fig:cqd-compare}a). Their experiments show that the simple classifier generalizes better when provided with the outputs of the complex classifier during training. This is based on an idea pioneered by Bucil\v{a} \etal~\cite{Caruana06} in a technique called ``model compression'' where simple classifiers such as linear models were trained to match the predictions of a decision-tree ensemble, leading to compact models. Thus, \cqd can be seen as a generalization of model compression when the domains of the two classifiers ${\cal A}$ and ${\cal B}$ are different (Figure~\ref{fig:cqd-compare}d). 
``Learning without forgetting''~\cite{li2016learning} shows that applying distillation on transfer learning can outperform fine-tuning, and has similar performance with multitask learning (joint training) but without using the data of original task. In this paper, we focus on domain adaptation problem where the tasks are the same but with paired data from different domains.

\vspace{-0.1in}
\paragraph{Learning with privileged Information} The framework of learning with privileged information (LUPI)~\cite{LUPI} (Figure~\ref{fig:cqd-compare}b) deals with the case when additional information is available at training time but not at test time. The general idea is to use the side information to guide the training of the models. For example, the SVM+ approach~\cite{LUPI} modifies the margin for each training example using the side information to facilitate the training on the input features. 
Most of these approaches require an explicit representation of the side information, i.e., the domain \cA can be written as a combination of domain \cB and side information domain ${\cal S}$. 
For example, such models have been used to learn classifiers on images when additional information about them such as tags and attributes are available at training time.
We note that Lopez-Paz \etal~\cite{unify_distill} made a similar observation unifying distillation and learning with privileged information.
\vspace{-0.1in}
\paragraph{Domain adaptation} Techniques for domain adaptation addresses the performance loss due to domain-shift from training to testing, leading to degradation in performance. 
For example, visual classifiers trained on clutter-free images do not generalize well when applied to real-world images. 
Typically it is assumed that a large number of labeled examples exist for the source domain, but limited to no labeled data is available for the target domain. 
To increase feature generalization, some approaches~\cite{tzeng2014deep,long2015learning} minimize the domain discrepancy through Maximum Mean Discrepancy (MMD)~\cite{gretton2012kernel}.
Other approaches learn a parametric transformation to align the representations of the two domains~\cite{saenko2010adapting,kulis2011you, fernando2013unsupervised, sun2015return}.
Bousmalis \etal~\cite{bousmalis2016domain} combines encoder-decoder structure and different loss functions to learn shared and domain-specific features explicitly.
Ganin \etal~\cite{ganin2016domain} proposed the domain-adversarial neural networks (DANN) which learns representations by competing with an adversarial network trained to discriminate the domains. 
Instead of learning domain-invariant features, some approaches~\cite{liu2016coupled, bousmalis2017unsupervised, Adda_CVPR2017} use Generative Adversarial Networks (GANs) to generate images of target domain for unsupervised domain adaptation.

When some labeled data is available for the target domain (supervised case), methods for multi-task learning~\cite{caruana1997multitask} are also applicable, including ones that are ``frustratingly easy''~\cite{daume2007frustratingly}. \cqd is a special case of supervised domain adaptation where we have correspondence between samples from the source and target domain, i.e., in supervised domain adaptation we have training data of the form $({\bf x}_i, y_i), {\bf x}_i \in {\cal A}$ and $({\bf z}_j, y_j), {\bf z}_j \in {\cal B}$, where ${\bf x}_i$ and ${\bf z}_j$ are drawn independently from the source and target domain respectively, and $y_i, y_j \in {\cal Y}$. In \cqd we know that ${\bf x}_i$ and ${\bf z}_i$ are two views of the same instance. This provides richer information to adapt models across domains. Our experiments show that distillation leads to greater improvements in accuracy compared to fine-tuning, a commonly used approach for domain adaptation, and ``staged training''~\cite{Saenko_ICIP16}, specifically designed for scenarios like ours where high-quality data is available at training time. 
The idea of transferring task knowledge through distillation has been applied for simultaneous domain adaptation and task transfer by Tzeng \etal~\cite{Tzeng_2015_ICCV}. They tried to match the average predicted label scores across examples in source domain to that of the target domain as instances lack one-to-one correspondence. 
In contrast, paired data in \cqd allows matching of label distributions on per-instance basis.

\vspace{-0.1in}
\paragraph{Cross modal learning} When multiple modalities of images are simultaneously available, the information about one domain can guide representation learning for another domain. Recent works have used a similar idea to ours to learn representations of depth from color using RGB-D data~\cite{Gupta_CVPR2016,Hoffman_ICRA2016},  representations of video from ambient sound~\cite{owens2016ambient} and vice-versa~\cite{aytar2016soundnet}, as well as visual representations through self-supervised colorization~\cite{larsson2016learning,zhang2016colorful}.
Our work identifies several novel situations when distillation can be applied effectively. For example, we train a model to recognize distorted images of birds by distilling a model trained on non-distorted ones.

\section{CQD Framework}\label{section:framework}
Assume that we have data in the form of $({\bf x}_i, {\bf z}_i,y_i)$, $i = 1, 2, \ldots, n$ where ${\bf x}_i \in {\cal A} $ is the high-quality data, ${\bf z}_i \in {\cal B}$ is the corresponding low-quality data, and $y_i \in {\cal Y}$ is the target label. In practice only the high-quality data ${\bf x}_i$ is needed since ${\bf z}_i$ can be generated from ${\bf x}_i$. The idea of \cqd is to first train a model $f$ to predict the labels on the high-quality data and train a second model $g$ on the low-quality data by forcing an agreement between their corresponding predictions by minimizing the following objective (Fig.~\ref{fig:cqd-framework}):
\begin{equation}\label{eqn:cqd}
 \sum_{i=1}^{n} {\cal L}_1\left( g({\bf z}_i), y_i \right) + \lambda \sum_{i=1}^{n} {\cal L}_2\left(g({\bf z}_i), f({\bf x}_i)\right) + {\cal R}(g).
\end{equation}

Here, ${\cal L}_1$ and ${\cal L}_2$ are loss functions, $\lambda$ is a trade-off parameter, and ${\cal R}(g)$ is a regularization term. The intuition for this objective is that by imitating the prediction of $f$ on the high-quality data $g$ can learn to generalize better on the low-quality data.

All our experiments are on multi-class classification datasets and we model both $f$ and $g$ using multi-layer CNNs, pre-trained on ImageNet dataset, with a final softmax layer to produce class probabilities ${\bf p} = \sigma({\bf z})$, i.e., $p_k = e^{z_k}/\sum_j e^{z_j}$. We use the cross-entropy loss ${\cal L}_1 ({\bf p}, {\bf q}) = \sum_i q_i \log p_i$, and the cross-entropy of the predictions smoothed by a temperature parameter $T$ for ${\cal L}_2({\bf p}, {\bf q}) = {\cal L}_1 \left(\sigma\left(\log({\bf p})/T \right), \sigma\left(\log({\bf q})/T \right) \right)$. When $T = 1$, this reduces to the standard cross-entropy loss. We also found that squared-error between the logits ({\bf z}) worked similarly. More details can be found in the experiments section.

\section{Experiments}\label{section:experiments}
We begin by describing datasets, models, and training protocols used in our experiments. Section~\ref{sec:exp:cqd} describes the results of various experiments on \cqd. Section~\ref{sec:exp:simultaneous} describes experiments for simultaneous quality distillation and model compression. Finally, Section~\ref{sec:exp:analysis} visualizes the distilled models to provide an intuition of why and how distillation works.

\begin{figure*}[t]
\centering
\includegraphics[width=1.00\linewidth]{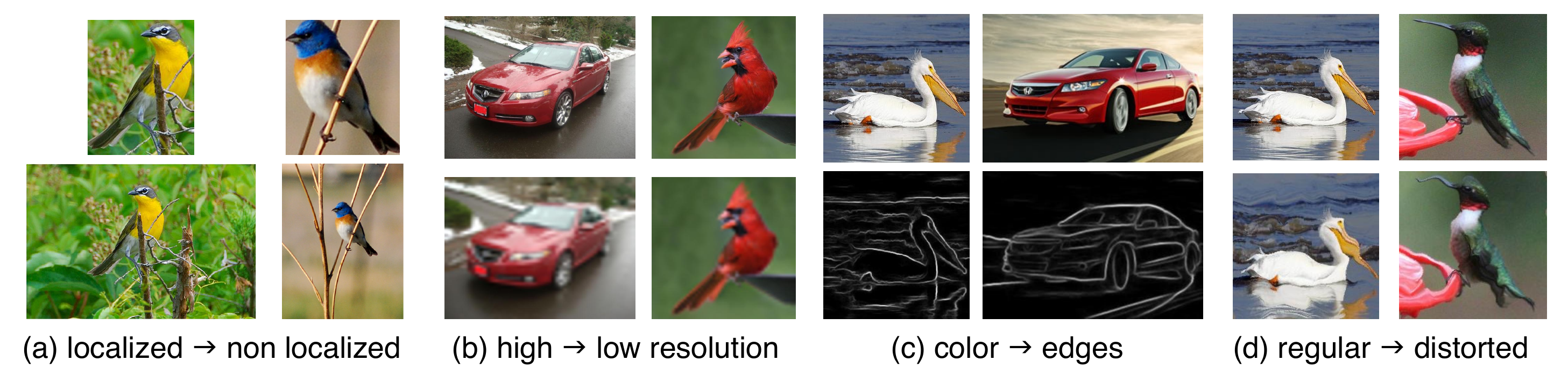} 
\vspace{0.02in}
\caption{\label{fig:dataset} Examples images from various cross-quality datasets used in our experiments. Images are from the birds~\cite{WahCUB_200_2011} and cars dataset~\cite{krause20133d}. In each panel, the top row shows examples of the high-quality images and the bottom row shows examples of the corresponding low-quality images. These include (a) localized and non-localized images, (b) high- and low-resolution images, (c) color and edge images, and (d) regular and distorted images.}
\vspace{-0.1in}
\end{figure*}

\paragraph{\textbf{Datasets}}
We perform experiments on the CUB 200-2011 dataset~\cite{WahCUB_200_2011} consisting of 11,788 images of 200 different bird species, and on the Stanford cars dataset~\cite{krause20133d} consisting of 16,185 images of 196 cars of different models and makes. Classification requires the ability to recognize fine-grained details which is impacted when the quality of the images is poor. Using the provided images and bounding-box annotations in these datasets, we create several cross-quality datasets which are described in detail in the Section~\ref{sec:exp:cqd} and visualized in Figure~\ref{fig:dataset}. We use the training and test splits provided in the datasets. 

\paragraph{\textbf{Models}} In our experiments, both $f$ and $g$ are based on CNNs pre-trained on the ImageNet dataset~\cite{deng09imagenet}. In particular we use \texttt{vgg-m}~\cite{chatfield14return} and \texttt{vgg-vd} models~\cite{simonyan14very} which obtain competitive performance on the ImageNet dataset.
While there are better performing models for these tasks, \eg those using novel model architectures\cite{lin2015bilinear,cimpoi15deep,sermanet14attention,jaderberg2015spatial}, and using additional annotations to train part and object detectors~\cite{bourdev2011describing,zhang14part-based,zhang14panda,branson14bird}, we perform experiments with simple models in the interest of a detailed analysis. However, we believe that our method is general and can be applied to other recognition architectures as well.

\paragraph{\textbf{Methods}} Below we describe various methods used in our experiments:
\begin{enumerate}[wide, labelwidth=!, labelindent=0pt, itemsep=0ex]
\item \textbf{Train on \cA:} \label{A} Starting from the ImageNet pre-trained model, we replace the 1000-way classifier (last layer) with a k-way classifier initialized randomly and then fine-tune the entire model with a small learning rate on domain \cA. This is a standard way of transfer learning using deep models, and has been successfully applied for a number of vision tasks including object detection, scene classification, semantic segmentation, texture recognition, and fine-grained classification~\cite{donahue13decaf,razavin14cnn-features,girshick14rich,cimpoi14describing,long2014fully,mostajabi2014feedforward,lin2015bilinear}.

\item \textbf{Train on \cB:} Here we fine-tune the ImageNet pre-trained model on domain \cB.

\item \textbf{Train on \cAB:} Here we fine-tune the model on domain \cA combined with domain \cB. Data augmentation is commonly used while training CNNs to make them more robust.

\item \textbf{Train on \cA, then train on \cB:} This is a combination of \cA and \cB where the fine-tuning on domain \cB is initialized from the model fine-tuned on domain \cA. This ``staged training'' was recently proposed in \cite{Saenko_ICIP16} as a state-of-the-art technique for low-resolution image recognition. However, this method can only be applied when both $f$ and $g$ have the same structure. This is denoted by \cAsB in our experiments.

\item \textbf{Cross quality distillation (CQD):} Here we use a model $f$ trained on domain \cA (Method~\ref{A}) to guide the learning of a second model $g$ on domain \cB using \cqd (Equation~\ref{eqn:cqd}). Like before, when $f$ and $g$ have identical structure we can initialize $g$ from $f$ instead of the ImageNet model with random weights for the last layer.
\end{enumerate}

\paragraph{Optimization details} There are two parameters, $T$ and $\lambda$, in the \cqd model. 
The optimal value we found on validation set is $T=10$ for all experiments, and $\lambda=200$ for the CUB, $\lambda=50$ for the CARS dataset.
The optimization in Equation~\ref{eqn:cqd} was solved using batch stochastic gradient descent, with learning rate starting from \texttt{0.0005}~(\texttt{0.0005} for CUB, \texttt{0.001} for CARS) changing linearly to \texttt{0.00005} after 30 epochs. Other parameters are as follows: \texttt{momentum=0.9, weight decay=0.0005, batch size=128} (\texttt{=32} when training \texttt{vgg-vd}). Instead of cross-entropy we also tried squared-distance on the logits {\bf z} as the loss function~\cite{ba2014deep}. There was no significant difference between the two and we used cross-entropy for all our experiments. Our implementation is based on MatConvNet~\cite{vedaldi14matconvnet-convolutional}.

\subsection{Cross Quality Distillation Results}
We experiment with five different kinds of quality reduction to test the versatility of the approach. For each case we 
report per-image accuracy on the test set provided in the dataset. Results using the \texttt{vgg-m} model for both function $f$ and $g$ are summarized in Table~\ref{tab:main-results} and are described in detail below. The main conclusions are summarized in the end of this section.

\label{sec:exp:cqd}
\newcolumntype{D}{>{\centering\arraybackslash}p{5.0em}}
\newcolumntype{C}{>{\centering\arraybackslash}p{3.5em}}
\begin{table}[t]
\begin{center}
\footnotesize{
\begin{tabular}{c|c|c|c|*{4}{c|}c|c}
\multirow{2}{*}{Description} & \multirow{2}{*}{Method} & \multirow{2}{*}{Test}& \multicolumn{1}{c|}{Local.} & \multicolumn{2}{c|}{Resolution}  & \multicolumn{2}{c|}{Edge} & \multicolumn{1}{c|}{Dist.} & Local. + Res.\\ 
\cline{4-10} 
 &  &  & CUB & CUB & CARS & CUB & CARS & CUB & CUB\\ 
\hline 
\rowcolor{Gray}
Upper bound & \cA & \cA & 67.0 & 67.0 & 59.3 & 67.0 & 59.3 & 67.0 & 67.0\\ 
No adaptation & \cA & \cB & 57.4 & 39.4 & 7.6 & 1.9 & 4.2 & 49.7 & 24.9\\ 
Fine-tuning & \cB & \cB & 60.8 & 61.0 & 41.6 & 29.2 & 45.5 & 58.4 & 46.2\\ 
Data augment. & \cAB & \cB &63.6  & 62.2  & 47.3 & 32.5 & 51.3 & 61.7 & 51.7\\
Staged training & \cA,\cB & \cB & 62.4 & 62.3 & 48.4 & 30.4 & 50.1 & 60.9 & 50.4\\ 
Proposed & \cqd & \cB & \textbf{64.4} & \textbf{64.4} & \textbf{48.8} & \textbf{34.1} & \textbf{51.5} & \textbf{63.0} & \textbf{52.7}\\ 
\end{tabular}
}
\end{center}
\caption{\label{tab:main-results}
\textbf{Cross quality distillation results.} Per-image accuracy on birds dataset (CUB)~\cite{WahCUB_200_2011} and Stanford cars dataset (CARS)~\cite{krause20133d} for various methods and quality losses. All results are using $f=g=\texttt{vgg-m}$ model. Training on \cA and testing on \cA is the upper bound of the performance in each setting (top row). Training on \cA and testing on \cB (no adaptation) often leads to a significant loss in performance. The proposed technique (\cqd) outperforms fine-tuning (\cB), data augmentation (\cA+ \cB), and staged training (\cA,\cB)~\cite{Saenko_ICIP16} on all datasets.}
\end{table}

\subsubsection{Localized to Non-localized Distillation}
To create the high-quality data, we use the provided bounding-boxes in the CUB dataset to crop the object in each image. In this dataset, birds appear in various locations and scales and in clutter. 
Therefore, \texttt{vgg-m} trained and evaluated on the localized data obtains 67.0\% accuracy, but when applied the non-localized data obtains only 57.4\% accuracy (Table~\ref{tab:main-results}). When the model is trained on the non-localized data the performance improves to 60.8\%. Staged training \cA,\cB improves the performance to 62.4\%, but \cqd improves further to 64.4\%.

For this task another baseline would be to train an object detector which first localizes the objects in images. For example, Krause \etal~\cite{krause2015fine} report around 2.6\% drop in accuracy (67.9\% $\rightarrow$ 65.3\%) when a R-CNN based object detector is used to estimate bounding-boxes of objects at test time instead of using true bounding-boxes (Table 2 in~\cite{krause2015fine}, CNN+GT BBox+ft vs. R-CNN+ft). Remarkably, using \cqd~we observe only 2.6\% drop in performance (67.0\% $\rightarrow$ 64.4\%) without running any object detector. Moreover, our method only requires a single CNN evaluation and hence is faster. In Section~\ref{sec:exp:analysis} we provide insights into why the distilled model performs better on non-localized images.

\subsubsection{High to Low Resolution Distillation}
Here we evaluate how models perform on images of various resolutions. For the CUB dataset we use the localized images resized to $224\times 224$ for the high-resolution images, downsample to $50\times 50$, and upsample to $224\times224$ again for the low-resolution images. For the CARS dataset we do the same but for the entire image (bounding-boxes are not used).

The domain shift leads to large loss in performance here. On CUB the performance of the model trained on high-resolution data goes down from 67.0\% to 39.4\%, while the performance loss on CARS is even more dramatic going from 59.3\% to a mere 7.6\%. Man-made objects like cars contain high-frequency details such as brand logos, shapes of head-lights, etc., which are hard to distinguish in the low-resolution images. A model trained on the low-resolution images does much better, achieving 61.0\% and 41.6\% accuracy on birds and cars respectively. Color cues in the low-resolution are much more useful for distinguishing birds than cars which might explain the better performance on birds. Using \cqd the performance improves further to 64.4\% and 48.8\% on the low-resolution data. On CARS the effect of both staged training and \cqd is significant, leading to more than 7\% boost in performance.


\vspace{-0.1in}
\subsubsection{Color to Edges Distillation}
Recognizing line-drawings can be used for retrieval of images and 3D shapes using sketches and has several applications in search and retrieval. As a proxy for line-drawings, we test the performance of various methods on edge images obtained by running the structured edge detector~\cite{Dollar13} on the color images. In contrast to low-resolution images, edge images contain no color information but preserve most of the high-frequency details. This is reflected in the better performance of the models on CARS than CUB dataset (Table~\ref{tab:main-results}). Due to the larger domain shift, a model trained on color images performs poorly on edge images, obtaining 1.9\% and 4.2\% accuracy on CUB and CARS receptively.

Using \cqd the performance improves significantly from 45.5\% to 51.5\% on CARS. On the CUB dataset the performance also improves from 29.2\% to 34.1\%. The strong improvements on recognizing line drawings using distillation and staged training suggests that a better strategy to recognize line drawings of shapes used in various sketch-based retrieval applications~\cite{su2015multi, wang15sketch} is 
to first fine-tune the model on realistically rendered 3D models (\eg with shading and texture) then distill the model to edge images.

\vspace{-0.1in}
\subsubsection{Non-distorted to Distorted Distillation} 
Here the high-quality dataset is the localized bird images. To distort an image as seen in Figure~\ref{fig:dataset}d, we use the thin plate spline transformation with uniform grid of 14$\times$14 control points. Each control point is mapped from a regular grid to a point randomly shifted by Gaussian distribution with zero mean and 4 pixels variance. 
Recognizing distorted images is challenging, and the performance of a model trained and evaluated on such images is 8.6\% worse (67.0\% $\rightarrow$ 58.4\%). Using \cqd the performance improves from 58.4\% to 63.0\%. 

On this dataset a baseline would be to remove the distortion by alignment methods such as congealing~\cite{congealing12}, or use a model that estimates deformations during learning, such as spatial transformer networks~\cite{jaderberg2015spatial}. These methods are likely to work well but they require the knowledge of the space of transformations and are non-trivial to implement. On the other hand, \cqd is able to nearly halve the drop in performance of the same CNN model without any knowledge of the nature of distortion and is easy to implement. Thus, \cqd may be used whenever we can model the distortions algorithmically. For example, computer graphics techniques can be used to model the distortions from underwater imaging.

\subsubsection{Color to Non-localized and Low Resolution Distillation}
Here the images has two different degradations at the same time: the low-quality data is low resolution images with the object in clutter, where the high-quality data is high resolution images cropped by the bounding boxes provided in the CUB dataset. 
Without adaptation, the performance drops to 24.9\%, more than when only have one type of degradation (57.4\% and 39.4\% separately).
We want to stress that the type of degradation in domain \cB can be arbitrary, as long as we have the instance-level correspondence between different domains which can be done by applying known transformations. 
As shown in the last column of Table~\ref{tab:main-results}, \cqd improves 6.5\% (46.2\% $\rightarrow$ 52.7\%) over fine-tuning.

\paragraph{Summary}
In summary we found that domain adaptation is critical since the performance of models trained on high-quality data is poor on the low-quality data.
Data augmentation (\cAB) and staged training (\cAsB) are quite effective, but \cqd provides better improvements suggesting that adapting models on a per-example basis improves knowledge transfer across domains.
\cqd is robust and only requires setting a handful of parameters, such as $T$ and $\lambda$, across a wide variety of quality losses. In most cases, \cqd cuts the performance gap between the high- and low-quality data in half.


\subsection{Simultaneous \cqd and Model Compression}\label{sec:exp:simultaneous}
In this section we experiment if a deeper CNN trained on high-quality data can be distilled to a shallow CNN trained on the low-quality data. This is the most general version of \cqd where both the domains and functions $f,g$ change. The formulation in Equation~\ref{eqn:cqd} does not require $f$ and $g$ to be identical. However, \cAB and \cAsB baselines cannot be applied here.

We perform experiments on the CUB dataset using localized and non-localized images described earlier. The deeper CNN is the sixteen-layer ``very deep'' model (\texttt{vgg-vd}) and the shallow CNN is the five-layer \texttt{vgg-m} model used in the experiments so far. The optimal parameters obtained on the validation set for this setting were $T=10, \lambda = 50$.

\begin{wraptable}{r}{0.59\textwidth}
\vspace{-0.3in}
\begin{center}
\small{
\renewcommand\arraystretch{1.1}
\begin{tabular}{c|c|c|c}
&  \multicolumn{3}{c}{training $\rightarrow$ testing}\\
\cline{2-4} 
$f \rightarrow g$  & \cA $\rightarrow$ \cA & \cB $\rightarrow$ \cB  &  \cqd $\rightarrow$ \cB\\ 
\hline 
\texttt{vgg-m} $\rightarrow$ \texttt{vgg-m}  & 67.0 & 60.8 & 63.7\\
\texttt{vgg-vd} $\rightarrow$ \texttt{vgg-m~} & - & - & 64.6\\
\texttt{vgg-vd} $\rightarrow$ \texttt{vgg-vd} & 74.9 & 69.5 & 72.4 \\
\end{tabular}
}
\end{center}
\caption{\label{tab:simultaneous-distillation} Accuracy of various techniques on the CUB localized/non-localized dataset.}
\vspace{-0.17in}
\end{wraptable}

The results are shown in Table~\ref{tab:simultaneous-distillation}. The first row contains results using \cqd~for \texttt{vgg-m} model which are copied from Table~\ref{tab:main-results} for ease of comparison. The third row shows the same results using the \texttt{vgg-vd} model. The accuracy is higher across all tasks. \cqd leads to an improvement of 2.9\% (69.5\% $\rightarrow$ 72.4\%) for the deeper model. The middle row shows results for training the \texttt{vgg-m} model on non-localized images from a
\texttt{vgg-vd} model trained on the localized images. This leads to a further improvement of 0.9\% (63.7\% $\rightarrow$ 64.6\%) suggesting that model compression and cross quality distillation can be seamlessly combined.

\section{Understanding Why \cqd Works}\label{sec:exp:analysis}

\paragraph{Relation to curriculum learning} Curriculum learning is the idea that models generalize better when training examples are presented in the order of their difficulty. Bengio \etal~\cite{bengio2009curriculum} showed a variety of experiments where non-convex learners reach better optima when more difficult examples are introduced gradually. 
In one experiment a neural network was trained to recognize shapes. There were two kinds of shapes: \texttt{BasicShapes} which are canonical circles, squares, and triangles, and \texttt{GeomShapes} which are affine distortions of the \texttt{BasicShapes} on more complex backgrounds. 
When evaluated only on test set of \texttt{GeomShapes}, the model first trained on \texttt{BasicShapes} then fine-tuned on \texttt{GeomShapes}, performed better than the model only trained on \texttt{GeomShapes}, or the one trained with a random ordering of both types of shapes.

We observe a similar phenomenon when training CNNs on low-quality data. For example, on the CARS dataset, staged training~\cite{Saenko_ICIP16} \cA,\cB outperforms the model trained on low-resolution data \cB, when evaluated on the low-resolution data \cB (48.4\% vs.~41.6\%). Since low-quality data is more difficult to recognize, introducing them gradually might explain the better performance of the staged training and \cqd techniques. Additional benefits of \cqd come from the fact that paired high- and low-quality images allowing better knowledge transfer through distillation.

\begin{figure}[t]
\centering
\begin{tabular}{cc}
\begin{adjustbox}{valign=t}
\begin{tabular}{@{}c@{}}
\includegraphics[width=0.16\linewidth]{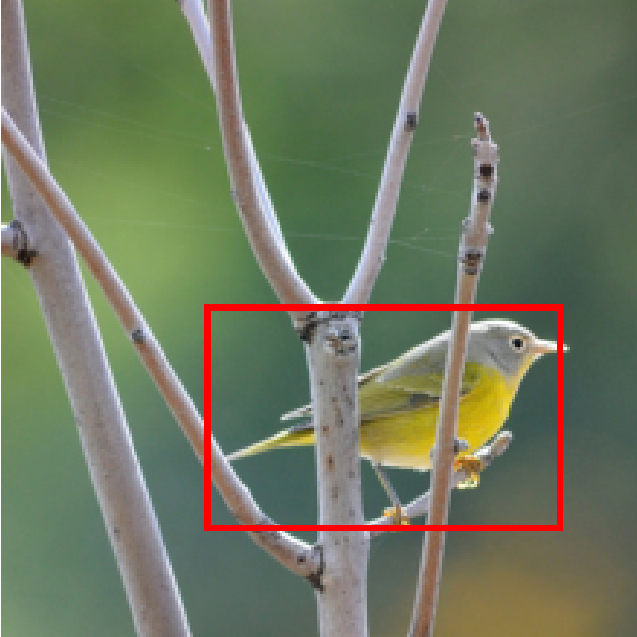} 
\includegraphics[width=0.16\linewidth]{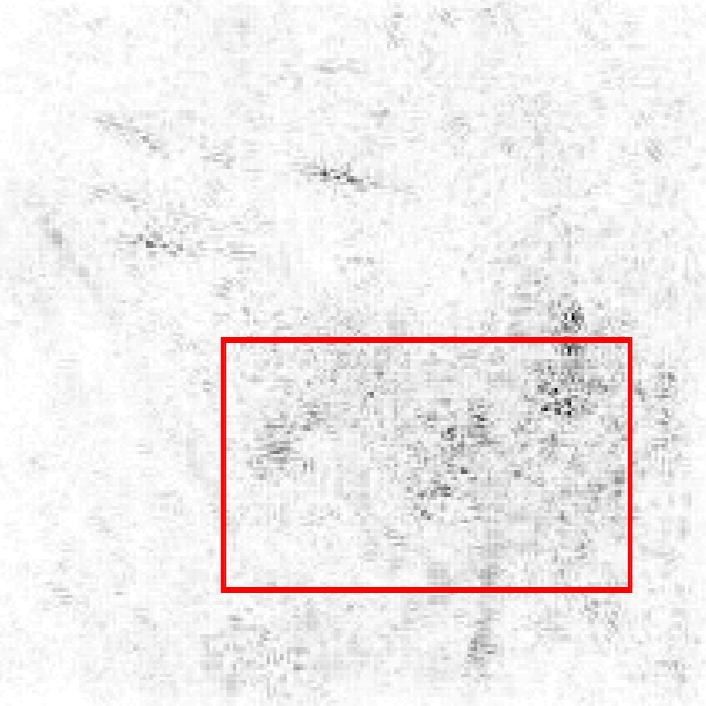} 
\includegraphics[width=0.16\linewidth]{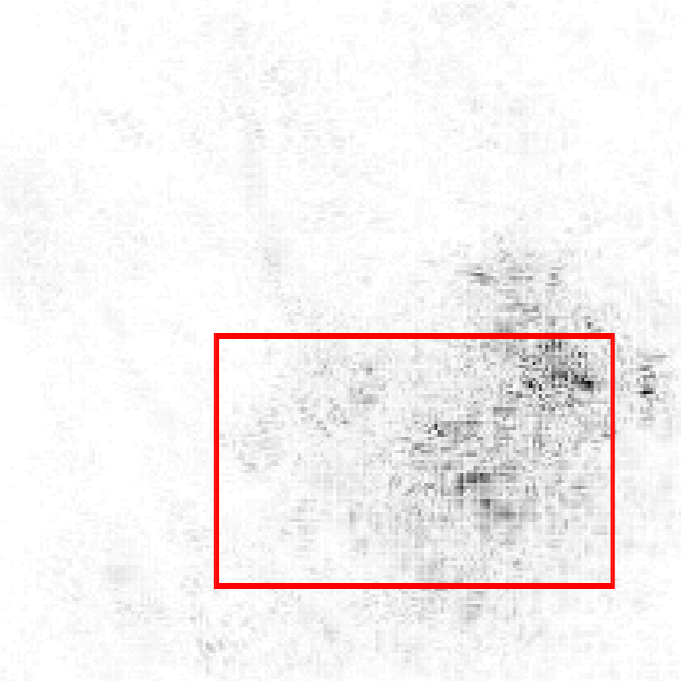} \\
\includegraphics[width=0.16\linewidth]{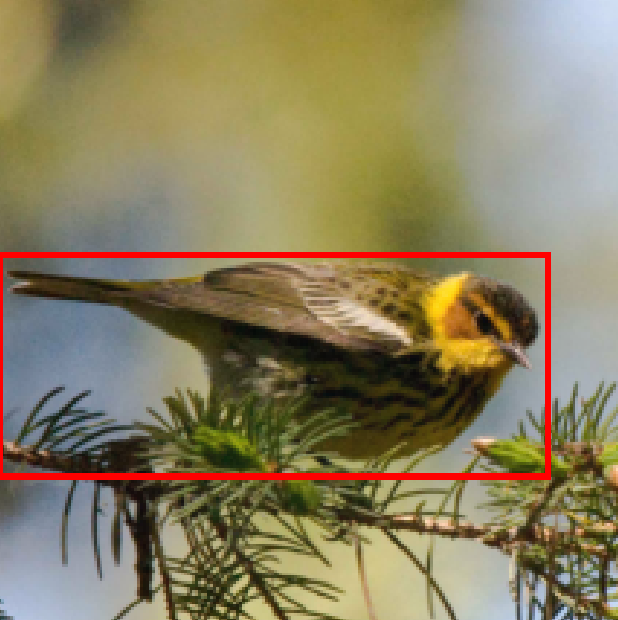} 
\includegraphics[width=0.16\linewidth]{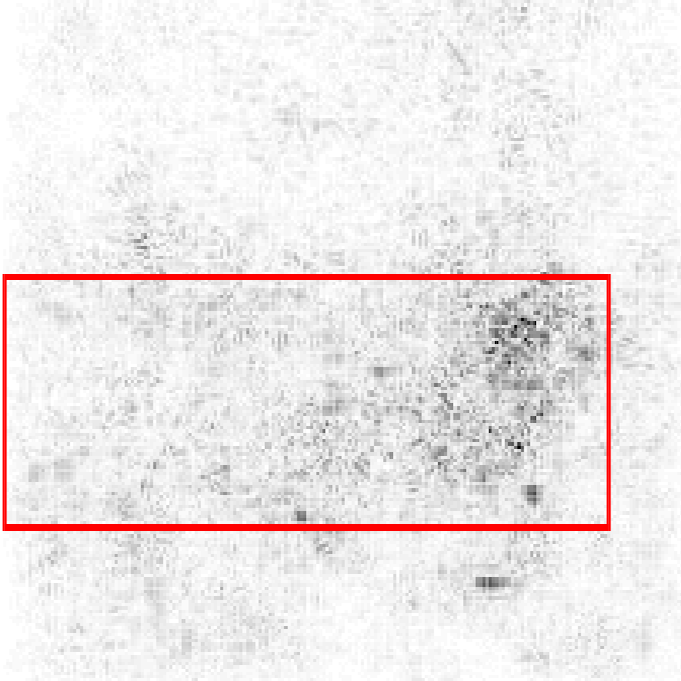} 
\includegraphics[width=0.16\linewidth]{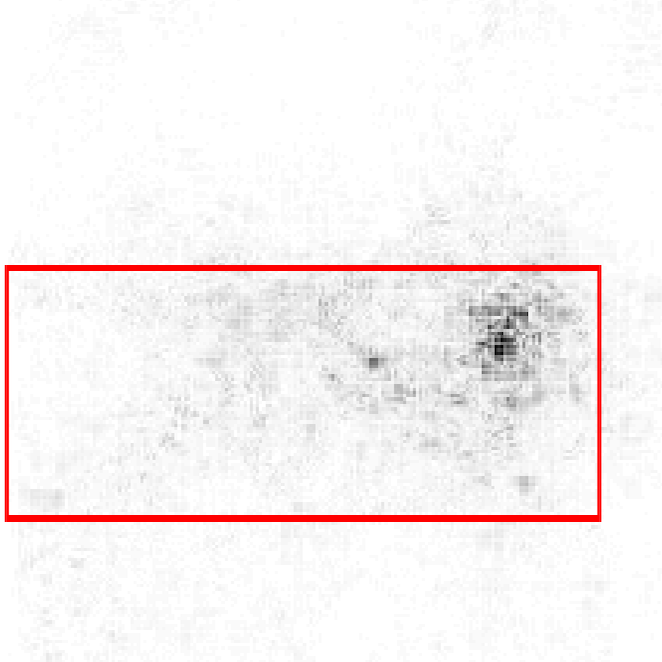} \\
\hspace{2.0em} image \hspace{2.0em} model = \cB \hspace{0.5em} model = \cqd \hspace{0.5em}
\end{tabular}
\end{adjustbox}
&
\begin{adjustbox}{valign=t}
  \includegraphics[width=0.38\linewidth]{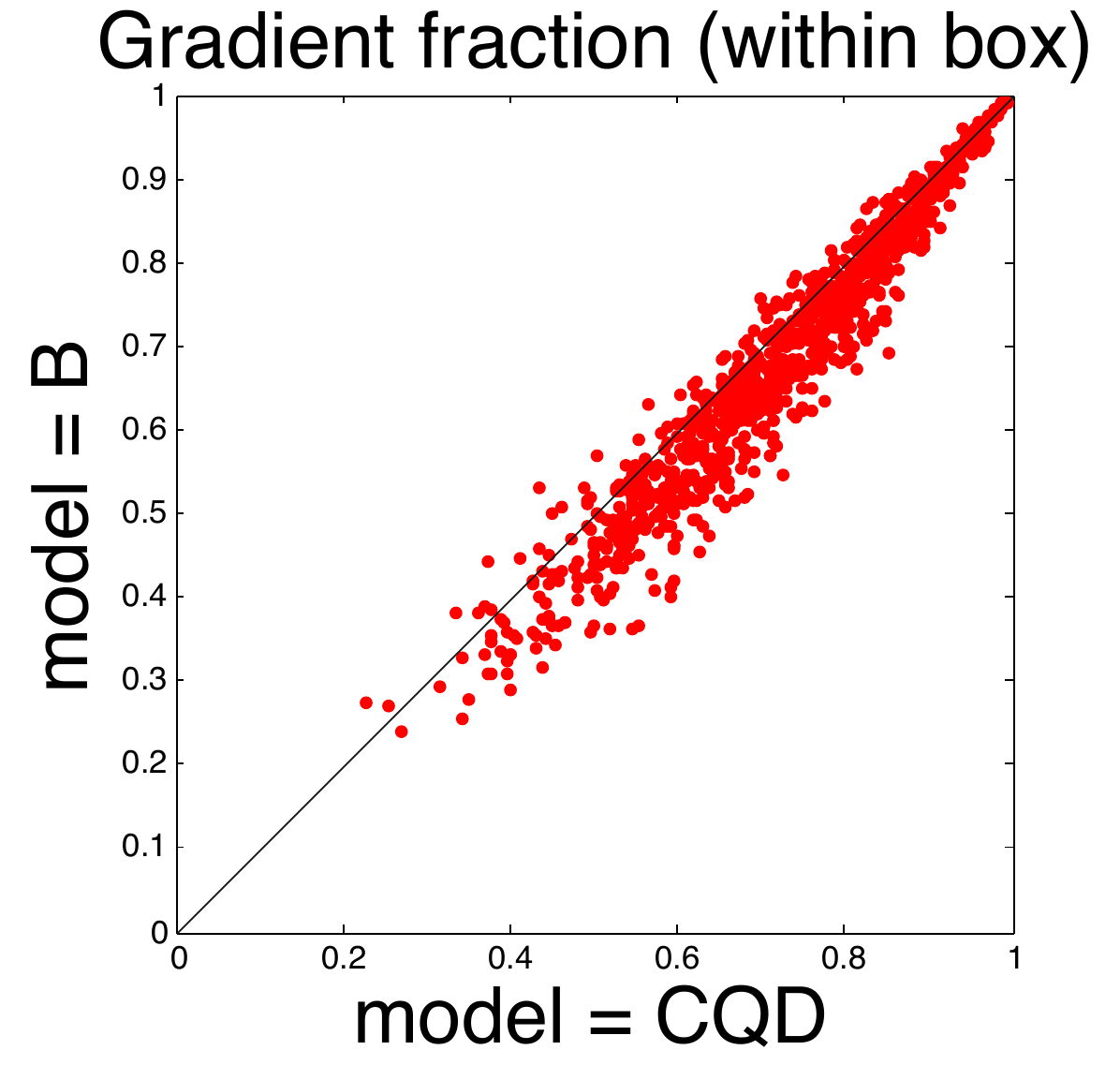} 
\end{adjustbox}
\end{tabular}
\vspace{0.15in}
\caption{\textbf{Left:} Image and gradient of the image with respect to the true class label for the model trained on \cB (non-localized images) and \cqd (from a model trained on localized images). Darker pixels represent higher gradient value. The gradients of the model trained using \cqd are more focused on the foreground object. \textbf{Right:} The scatter plot of the fraction of total gradient within the bounding-box for 1000 training images for the two models. 
}
\label{fig:example}
\vspace{-0.15in}
\end{figure}

\paragraph{Understanding \cqd through gradient visualizations} 
Here we investigate how the knowledge transfer occurs between a model trained on localized images and non-localized images. Our intuition is that by trying to mimic the model trained on the localized images a model must learn to ignore the background clutter. In order to verify this, we compute the 
gradient of log-likelihood of the true label of an image with respect to the image using the \cqd model and \cB model, both are trained only on non-localized images. Figure~\ref{fig:example}-left shows the gradients for two different images. The darkness of each pixel $i$ is proportional to the norm of the gradient vector at that pixel $||G_i||_2, ~G_i = [G^r_i,G^g_i,G^b_i]$ for $r,g,b$ color channels. The gradients of the \cqd model are more contained within the bounding-box of the object, suggesting a better invariance to background clutter. As a further investigation we compute the fraction of gradients within the box:
$ 
\tau = ({\sum_{i \in \text{box}}||G_i||_2}) ~/~ ({\sum_{i \in \text{image}}||G_i||_2})
$.
This ratio is a measure of how localized the relevant features are within the bounding-box. A model based on a perfect object detector will have $\tau=1$. We compute $\tau$ for 1000 images for both \cqd and \cB models and visualize them on a scatter plot as seen in Figure~\ref{fig:example}-right. On average the \cqd model has higher $\tau$ than \cB model, confirming our intuition that the \cqd model is implicitly able to localize objects. 

\section{Conclusion}
We proposed a simple generalization of distillation, originally used for model compression, for cross quality model adaptation. We showed that \cqd achieves superior performance than domain adaption techniques such as fine-tuning on a range of tasks, including recognizing low-resolution images, non-localized images, edge images, and distorted images. Our experiments suggest that recognizing low-quality data is a challenge, but by developing better techniques for domain adaptation one can significantly reduce the performance gap between the high- and low-quality data. We presented insights into why \cqd works by relating it to various areas in machine learning and by visualizing the learned models.

Training highly expressive models with limited training data is challenging. A common strategy is to provide additional annotations to create intermediate tasks that can be easily solved. For example, annotations can be used to train part detectors to obtain pose, viewpoint, and location-invariant representations, making the fine-grained recognition problem easier. However, these annotation-specific solutions do not scale as new types of annotations become available. An alternate strategy is to use \cqd by simply treating these annotations as additional features, learning a classifier in the combined space of images and annotations, and then distilling it to a model trained on images only. This strategy is much more scalable and can be easily applied as new forms of side information, such as additional modalities and annotations, become available over time. In future work, we aim to develop strategies for distilling deep models trained from richly-annotated training data for better generalization from small training sets.

\paragraph{Acknowledgement:}
This research was supported in part by the NSF grants IIS-1617917 and ABI-1661259, and a faculty gift from Facebook. The experiments were performed using high performance 
computing equipment obtained under a grant from the Collaborative R\&D Fund 
managed by the Massachusetts Tech Collaborative and GPUs donated by NVIDIA.

\bibliography{egbib,bibliography/egbib,bibliography/publications,bibliography/bcnn,bibliography/bibliography}

\end{document}